# Multiple Waypoint Navigation in Unknown Indoor Environments

Shivam
Sood *
IIT Kharagpur
Kharagpur, India
0000-0001-8188-2065

Jaskaran
Singh Sodhi *
IIT Kharagpur
Kharagpur, India
0000-0003-2028-1439

Parv
Maheshwari *
IIT Kharagpur
Kharagpur, India
0000-0002-5330-3027

Karan
Uppal *
IIT Kharagpur
Kharagpur, India
0000-0001-5530-6179

Debashish
Chakravarty
IIT Kharagpur
Kharagpur, India
0000-0002-1247-8963

*Abstract*— Indoor motion planning focuses on solving the problem of navigating an agent through a cluttered environment. To date, quite a lot of work has been done in this field, but these methods often fail to find the optimal balance between computationally inexpensive online path planning, and optimality of the path. Along with this, these works often prove optimality for single-start single-goal worlds. To address these challenges, we present a multiple waypoint path planner and controller stack for navigation in unknown indoor environments where waypoints include the goal along with the intermediary points that the robot must traverse before reaching the goal. Our approach makes use of a global planner (to find the next best waypoint at any instant), a local planner (to plan the path to a specific waypoint) and an adaptive Model Predictive Control strategy (for robust system control and faster maneuvers). We evaluate our algorithm on a set of randomly generated obstacle maps, intermediate waypoints and start-goal pairs, with results indicating significant reduction in computational costs, with high accuracies and robust control.

*Keywords—Multiple Waypoint Navigation, Adaptive Model Predictive Control, Integrated Planning and Control, Sensor-Based Reactive Planning*

I. INTRODUCTION

The aim of the indoor motion planning problem is to find a path planning, control and mapping solution for a robot which can sense its local environment, and navigate safely through a set of rigid obstacles.

The planning module answers the question '*Where do I go?*' and aims to find a path for the agent to traverse which minimises a given constraint. The constraint could be a cost function (or budget) like battery, distance or time-to-goal. Existing work in planners can be primarily divided into three domains - discrete planners, sampling-based planners and learning-based planners. Discrete planners assume a discrete world and expand into the search-space sequentially (expand the tree by recursively exploring neighboring points). On the other hand, sampling-based planners like [1,2], plan by growing a tree in a continuous search-space and are optimal for high-dimensional planning. Learning-based planners like [3] are preferred in scenarios where pattern recognition is feasible and informed sampling can be of high significance.

The controls module needs to follow the given trajectory as closely as possible but at the same time the control actions should not be too aggressive as that would lead to wear and tear of the robot mechanism. Generally deployed controllers for these scenarios like PIDs (Proportional-Integral-Derivative)[4] do not have the ability to predict the sudden changes in the path and plan accordingly which is a highly desired trait for reactive planning and controls. Optimal predictive controls provide an interesting solution to the above problem with more intuitive tuning parameters. However really sharp turns might still cause a few unwanted behaviours similar to the case of extra smooth turns as both can lead to collisions. We present corrective turn algorithm for such cases and a simple adaptive weight strategy which makes it easier to tune the controller and inculcate the desired behaviours through a simple cost function.

Indoor navigation robots have vast area of applications, including social navigation [5], manipulation [6], GPS- denied navigation [7], automated cleaning and nuclear plant management [8]. These applications encounter both static and dynamic obstacles, which pose real-time challenges to mapping and path planning. In this work however, we explore the static obstacles case (due to the test worlds provided by the International Conference on Intelligent Robots and Systems (IROS) organizers).

A common sub-problem of indoor navigation is path planning and motion control in unknown environments by performing Simultaneous Localisation and Mapping (SLAM), while trying to maintain near-optimal planning. In this a crucial part is that of real-time reactive planning, obstacle avoidance and controls. This introduces a need to use a real-time updated obstacle map which is usually achieved using a depth-based sensor and/or a LiDAR. Global path planning in such cases is performed using online planners, which are dependent on an incrementally updated map of the environment. On the other hand, local planning and obstacle avoidance can be optimised for static or dynamic obstacles by studying the nature of the environment, indoor or off-road, sparse or cluttered, etc.

In this paper, we propose a planning and controls stack for a differential-drive mobile robot, for multiple waypoints navigation in unknown indoor environments. We have further extended the real-time application of our stack by using a novel probabilistic path-planner integrated with multiple-shooting based model predictive control (MPC) to robustly navigate a world with multiple obstacles in the shortest time possible.

---

[1] https://github.com/thisisjaskaran/multi-waypoint-indoor-navigation
[2] https://youtu.be/JAr6UzAz4KQ
*Equally contributing authors

This approach was used to develop the winning solution for the IROS-RSJ Navigation and Manipulation Challenge for Young Students 2021. Our code is available here[1]. The robot can be seen performing here[2].

## II. EXISTING METHODOLOGIES

This section presents a brief overview of the past work done in the field of indoor planning and controls. As explained before, the problem mainly aims to solve accurate, safe and fast indoor path planning and controls, while sometimes on a constrained budget.

State-of-the-art path planning algorithms, are capable of performing on-board efficient navigation and system control. Discrete planners like the Dijkstra Algorithm and A* planners have been explored for their applications in low-dimensional navigation stacks and discretized search spaces, owing to optimality and computational speed. Sampling-based planners on the other hand like the Dynamic Window Approach (DWA) [9] and a Frenet-frame based planner [10] have worked well in continuous environments, even considering that they require higher computation than discrete planners. Other sampling-based planners like Rapidly-exploring Random Trees (RRTs), and its variants like RRT* [1] and Informed-RRT* [2], are preferred in high-dimensional planning problems, or cases where the agent has higher degrees of freedom, and also promise asymptotic optimality. Works like [11] have also shown the efficient use of sampling-based algorithms like RRT to solve discretized space problems. These planners however, are used mostly in single-start single-goal environments in online planners. Bai, Tian and Fan in recent work [12] introduced a multiple-waypoint path planning method which employs the RRT* algorithm and adapts it to plan through multiple waypoints, by growing a tree from each waypoint. Maddi, Sheta and Mahdy in [13] exploited the search-space capability of genetic algorithms, combined with local search algorithms. Recently, a lot of work is also being done in the field of learning-based control-optimal path planning. These works can be broadly divided into Deep Learning and Reinforcement Learning based approaches. Some Deep Learning based planners like [3] and [14] explore how offline informed planning can be done for indoor environments given the encoding of an environment, along with the start and goal waypoints.

Other planners like [15] use a reward-based system to learn feasible paths, and have made advancements in the field of Reinforcement Learning based navigation. Learning-based methods however, suffer from high computational costs and often require heavy duty processing units and offline training. Even though sampling-based planners have often shown more promising results than discretised planners in most cases, the latter while using graphs to simulate the environment has been found to be optimal in real-time planning in low-dimensional and organised scenarios.

Differential Drive Mobile Robots (DDMR) are the most commonly used mobile robots with varied uses ranging from toys to factory and military environments. Thus, the control of such bots is a topic of wide interest where the majority of work has been done on PID and optimal control. Variations of PID controllers for this task have been widely explored. Steering control was studied by [4]. More robust control frameworks like pole-placement for linear feedback control were presented in [16], and then comes optimal control frameworks like Linear–quadratic regulator (LQR) [17] and MPC [18]. Works on MPC can further be divided into linear and non-linear MPC. The application of Linear Model Predictive Control strategy for mobile robots was studied by [19] while its stabilizing and regulating properties have been mentioned in [20]. A nonlinear MPC with Euler discretization for mobile robots was proposed by [21], where using multiple shooting accelerates the convergence of the optimization problem. Existing works also show the usage of adaptively changing the weights of the cost function to further improve the performance of MPC. An adaptive weight scheme based on the desired response was studied by [22] and a fuzzy adaptive weight MPC was presented by [23]. However, since these approaches change the weights at every iteration, these existing methods defy the purpose of warm setting as mentioned in Section IV-D. This makes the MPC solution converge slower since the initialization is no longer close to the actual solution. Our method provides simpler conditions for adaptive weights, changing the weights only when really necessary. These conditions also make it easy to tune while several other aiding optimizations help our controls module to perform better.

## III. PROBLEM FORMULATION

This problem is formulated according to the requirements of the competition, and the assumptions made are for the same. We have formulated our problem as a time-constrained navigation task for a differential bot in an unknown environment in which the agent state is defined by $(x, y, \theta)$ where $(x, y)$ are the positional coordinates and $\theta$ is the orientation yaw of the robot with respect to the world frame. A LiDAR was used to simulate indoor mapping, while Global Positioning System (GPS) and Inertial Measurement Unit (IMU) system were used to simulate indoor localisation. Each world consists of numerous waypoints which are always placed at centre of unit square space and hence the environment can be visualised as shown in Fig. 1. This does not mean that only grid-based movements are allowed to the robot, instead we leverage the grid-based formatting of the surroundings and waypoints to easily calculate global paths between the current position and the target waypoint. This is further explained in Section IV-B.

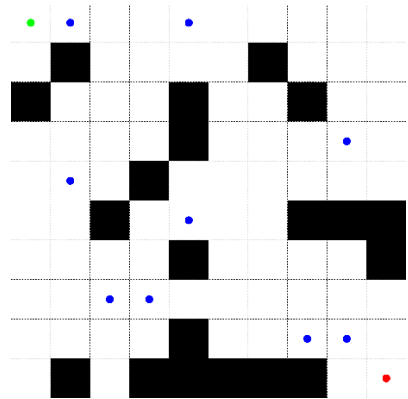

Fig. 1. Sample world with green - start position, red - goal position, blue - intermediate waypoints and black – obstacles

## IV. OUR APPROACH

We start by representing the environment in form of a undirected graph with orthogonal neighbours connected to each other. We use NetworkX [24] to build, update and query this graph. Distance between two nodes is defined as the minimal number of edges required to traverse while moving from the start point to the destination.

Then moving forward, a common approach in navigation tasks is to determine the next waypoint so as to optimise the targeted constraint and to then give a global path to the controls module which is responsible to track the global path while avoiding obstacles. This approach leads to an issue of higher computational requirements. This is because by adding obstacle avoidance to the cost function of the optimal controls module we are increasing the non-convexity of the optimisation space which requires a higher computation resources for real-time performance.

With accurate localisation and mapping (Section IV-A), to solve the previously mentioned issue, we have used an intermediary module between next best waypoint calculation (Section IV-B) and the controls module. This intermediary module is responsible for path planning (Section IV-C) whose responsibility is to generate an obstacle free path from the current state to the target waypoint.

As static obstacle avoidance is already considered in the global path provided by the path planning module to the controls module, we no longer require an obstacle avoidance term in the cost function inside our controls module. Because of this, the non-linear nature of the cost function of our controls module is decreased and hence the module can optimise the cost function much faster hence decreasing the real-time requirement of computational resources.

Finally, we introduce an adaptive-control stack in IV-F which provides user-dependent switches for weights, planning resolution and corrective turn.

### A. Localisation and Mapping

For accurate localisation and mapping, we use LiDAR-based incremental mapping fused with a GPS-IMU integrated localisation system which maintains a discrete map of the world. Since we showcase our approach in unknown environments in terms of positions of obstacles, we update the world graph as soon as we detect an obstacle to remove the corresponding vertex from the graph. This allows us to generate a obstacle free global path as the deleted vertices are not considered when the graph is queried to generate global path.

### B. Choosing the Next Waypoint

Calculating the global trajectory in an unknown environment requires an incrementally updating map which accounts for static obstacles. Due to this, the immediate next waypoint calculation for the agent becomes a real-time problem as the map contains uncertainty in future obstacles. Our novel path planner module tackles this problem by calculating global trajectories each time a static obstacle is detected and the map is updated, hence doing asynchronous planning for efficient real-time performance.

Let the list of waypoints left to be traversed be denoted as $P$ with $n$ as the number of elements in $P$. To calculate the next best waypoint $p_N$, we employ the following approaches:

*1) Greedy Approach:* This approach chooses the next target to be the waypoint which minimises the following cost function,

$$p_N = \min_{m \in P} J(m) = \sqrt{(x - x_m)^2 + (y - y_m)^2} \quad (1)$$

Here x and y are derived from agent's current state, while $x_m$ and $y_m$ represent the coordinates of $m$. This method, although sub-optimal, has a time complexity of $n$ and is thus suitable for online or real-time planning.

*2) Best-Cost Path (BCP) Approach:* In this, first current node of the robot according to the the world graph is inserted at the first position of $P$, and then this approach exploits all possible solutions ($W$) from the updated $P$. In all these permutations the first and the last element are fixed which are the current position and the final goal of the robot. An optimal permutation ($w_{opt}$) is decided by minimizing the sum of euclidean distances between consecutive chosen waypoints in that list. Then the second element of $w_{opt}$ is used as $p_N$. However, for a $P$ with $n$ waypoints left to traverse, this approach is computationally expensive with a time complexity of $(n!)$.

$$w_{opt} = \min_{w \in W} J(w) = \sum_{i=1}^{n-1} \sqrt{(x_{i+1} - x_i)^2 + (y_{i+1} - y_i)^2} \quad (2)$$

*3) Probabilistic Waypoint Approach:* A novel probablistic approach, which combines the online planning ability of the greedy method, and the optimality of the best-cost method, is introduced to perform near-optimal online planning. This approach performs a greedy search on a fractional search space (search-space limiting factor $\gamma$) generated by the permutations of best-cost path approach. This lowers the complexity of the solution by a factor of $\gamma$, with the tradeoff of always getting a near-optimal (if not optimal) cost. Detailed algorithm is given in Algorithm 1. Shuffling of the permutations is done to ensure a distribution of costs in each fractional search-space.

### C. Path Planning

In the first phase, given the next target waypoint and the current position, a raw path, which is a sequence of nodes to be traversed, is generated using NetworkX. Then in the next phase, the module uses a path resolution to granuarilise the raw path to make it finer which helps the controls module in better tracking of the path specially during the turns. Here we define resolution for path planning as number of points added in between the two consecutive points provided by the first stage of the planner. This planning module is responsible for adaptively changing the path resolution as explained in Section IV-E2.

Along with this, it is also responsible for fixing yaw of the path, applying turn correction and determining the best direction

**Algorithm 1** Probabilistic Waypoint Algorithm

```
 1: function EUCLIDIEANDIST(wp_list, first, second)
 2:     x₁, y₁ ← wp_list[first]
 3:     x₂, y₂ ← wp_list[second]
 4:     return √((x₂ − x₁)² + (y₂ − y₁)²)
 5: function DISTMETRIC(waypoints_list)
 6:     total_dist ← 0
 7:     for i ∈ range(len(waypoints_list)-1) do
 8:         total_dist += EuclideanDist(waypoints_list, i, i+1)
 9:     return total_dist
10: function FINDMIN(search_space)
11:     min_cost ← 0
12:     opt_perm ← []
13:     for perm ∈ search_space do
14:         cost = DistMetric(perm)
15:         if cost<min_cost then
16:             opt_perm = perm
17:             min_cost = cost
18:     return opt_perm
19: function PROBABILISTICWAYPOINT(current_waypoints, cur_node)
20:     \\ current_waypoints is a list of waypoints not traversed yet
21:     \\ cur_node is a tuple which denoted the node the robot
        currently is in according to the graph used in motion planning
        module
22:     waypoints = [cur_node] + current_waypoints
23:     all_paths ← list_permutations(waypoints)
24:     shuffle all_paths
25:     search_space_len ← int(len(all_paths) * γ)
26:     search_space ← all_paths[:search_space_len]
27:     best_permutation ← FindMin(search_space)
28:     chosen_waypoint = best_permutation[1]    ▷ Since the first
        element of best_permutation is the cur_node
29:     return chosen_waypoint
```

**Algorithm 2** Algorithm to fix yaw at discontinuities and take shorter turns

```
 1: if yaw_ref == π  and  yaw_cur < 0 then
 2:     yaw_cur ← yaw_cur + 2π
 3: else if yaw_ref == −π  and  yaw_cur > 0 then
 4:     yaw_cur ← yaw_cur − 2π
 5: if yaw_cur * yaw_ref < 0 then
 6:     if |yaw_cur − yaw_ref| > π then
 7:         if yaw_cur < 0 then
 8:             yaw_ref ← yaw_ref − 2π
 9:         else
10:             yaw_ref ← yaw_ref + 2π
```

for turning. The need for these responsibilities have been further explained.

While following a trajectory, if given the target yaw is among the points of discontinuity of tan(θ) i.e. {π, - π}, a little overshoot would cause the controller to become unstable as the current yaw might start oscillating above and below these values of discontinuity. Hence these conditions have to be especially considered as shown in Algorithm 2.

The need to consider the shortest turning direction also arises when the given target yaw is at the point of discontinuity of tan(θ). An example would be when the bot is at $yaw=π$ and you want it to move to $yaw = -π/2$. The controller not knowing about the discontinuity would move bot from yaw = π to yaw = 0 and then move to yaw = - π/2. The shorter turn behaviour is implemented for such cases as shown in Algorithm 2

Navigating in grid worlds often require you to take sharp turns. Although these behaviours can be implemented by over-tuning or changing the cost function, this would make the controller unstable in other situations like straight or diagonal paths. As an example, take an L shaped path. MPC's followed trajectory would depend on the prediction horizon and the cost function. A smooth turn might collide with the obstacles while an in place turn would require over tuning and more traversal time. A better workaround is our adaptive weight strategy (Section IV-E) used in addition with the corrective turns as shown in Fig. 4. This makes sure there is no collision with the obstacles based on a simple tuning parameter and the control behaviour specific to turns can be implemented without messing with the control behaviour in general.

### D. Motion Control

We use Model Predictive Control which is an advanced Optimal Control framework which optimizes the control action for $N$ time steps (called the prediction horizon). MPC does this by predicting the next $N$ states using a system model and iteratively optimizing the control values based on an objective cost function that consider constraints in its formulation, making it one of the few control algorithms that does this inherently. For the reference state values needed by MPC we use a KDTree search on the path to find the closest point on the path from the bot's position. However, because the system dynamics are simply approximate representations of the real world, the predicted trajectory may not always be followed entirely. To solve this problem MPC only implements the first control action, measures the new state and then re-optimizes the planned trajectory and the associated control vector based on that new state. As shown in [25], initializing the optimization variables in the next iteration with the previously computed estimates, also known as warm starting, reduces the computation since the (re)optimization now converges faster. MPC being predictive in nature presents substantial benefits over other control algorithms like geometric controllers that are myopic in nature. Formulating the control problem as a constrained optimization problem enables us to model high level task goals through simple cost functions and synthesize all the details of the behaviours of the control law automatically. Some major features of our controls module are –

*1) Control as an Optimization Problem:* The discrete time system dynamics of any system can be represented as:

$$x_{t+1} = f(x_t, u_t) \qquad (3)$$

where $x_t$ represents the state at time t, $u_t$ represents the control action to be taken at that time t and $f$ is the state dynamics.

Let $l(x_i, u_i)$ be the cost-to-go while at state $x_i$ and applying a control action $u_i$ and $l(x_N)$ be the cost of being at state $x_N$. Then the cost function $J$ is defined to be the sum of cost-to-go from the current and predicted states.

$$J(X, U) = \sum_{i=0}^{N-1} l(x_i, u_i) + l(x_N) \qquad (4)$$

where $X$ and $U$ are the vectors containing the $N+1$ states and the corresponding $N$ control actions. The optimal control problem (OCP) is framed as:

$$\min_{U} J(X, U)$$

$$s.t. \ x_{i+1} = f(x_i, u_i), \quad (5)$$

$$u_i \in \mathbb{U} \ \forall \ i \in [0, N-1],$$

$$\text{and } x_i \in \mathbb{X} \ \forall \ i \in [0, N]$$

where $\mathbb{U}$ and $\mathbb{X}$ represent the permitted values of the controls and states respectively. Here $x_0$ represents the bot's current position.

*2) OCP as a Non-Linear Programming Problem (NLP):* To solve our OCP we convert the problem into an NLP. We have used CasADi library [26] to solve our non-linear optimization problem.

*3) Single Shooting Approach:* We define the optimization variables as $W = [u_0, u_1 ... u_{N-1}]$. Equation (3) is used to get the next states as a function of our optimization variables.

$$X = \begin{bmatrix} x_0 \\ x_1 = f(x_0, u_0) \\ \vdots \\ x_N = f(f(x_{N-2}, u_{N-2}), u_{N-1}) \end{bmatrix} \quad (6)$$

The NLP is now formulated as:

$$\min_{W} J(X, W)$$

$$s.t. \ g(X, W) < 0 \quad (7)$$

where $g$ represents the equality and inequality constraints for the NLP. The actuator limits and the estate constraints are represented this way. Although single shooting implementation is straight forward and more intuitive to implement, it scales poorly with increasing N. Solving for the next states recursively makes the prediction calculation of the states highly non-linear as the prediction horizon is increased. We also cannot initialize the state trajectory with the previously predicted states, only the controls actions, as the states are not present in our optimization variables and so have to be calculated from scratch every time. Due to these drawbacks, we have used the multiple shooting approach to solving our MPC problem formulation.

*4) Multiple Shooting Approach:* The key idea behind multiple shooting (also known as lifted Single Shooting) MPC is to break down the prediction horizon time interval into shorter partitions and add the system model as a state constraint at each optimization step instead of solving for it recursively as shown in (9). So now, our optimization variables will also include the states.

$$W = [u_0, u_1 ... u_{N-1}, x_0, x_1 ... x_N] \quad (8)$$

Now the cost becomes a function of controls and the states ($W$) unlike in (7) where it was a recursive function of the optimization variables due to (6). This reformulation makes the NLP less non-linear and hence decreases the computation time. Also, the new NLP problem will have two sets of inequality and equality constraints, $g_1$ for the controls and state constraints and $g_2$ to make our predicted states equal to the actual states. The NLP can now be formulated as:

$$\min_{W} J(W)$$

$$s.t. \ g_1(f(W, x_0, T), W) \leq 0 \quad (9)$$

$$g_2 = \begin{bmatrix} \overline{x_0} - x_0 \\ f(x_0, u_0) - x_1 \\ \vdots \\ f(x_{N-1}, u_{N-1}) - x_N \end{bmatrix} = 0$$

*5) MPC for Differential Drive Mobile Robot:* Here we showcase how the above mentioned MPC algorithm can be deployed on a DDMR robot using a kinematic state space model.

Let $X_i$ and $U_i$ be the state and the control vector and $i^{th}$ time step.

$$X_i = [x_i, y_i, \theta_i]^T \quad U_i = [v_i, \omega_i]^T \quad (10)$$

where $x_i, y_i, \theta_i, v_i$ and $\omega_i$ represent the robot's $x$ and $y$ coordinates, yaw, linear velocity and angular velocity respectively for the $i^{th}$ timestep. All these are in the global frame as can be seen in Fig 2.

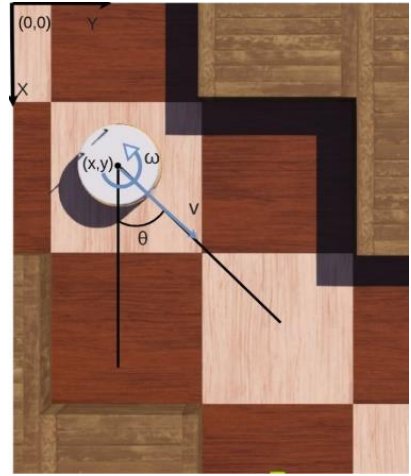

Fig. 2. State and control variables shown with respect to the fixed inertial coordinate system in the top left.

The kinematic state space model for the DDMR becomes:

$$\begin{bmatrix} \dot{x}_i \\ \dot{y}_i \\ \dot{\theta}_i \end{bmatrix} = \begin{bmatrix} \cos\theta_i & 0 \\ \sin\theta_i & 0 \\ 0 & 1 \end{bmatrix} \begin{bmatrix} v_i \\ \omega_i \end{bmatrix} \quad (11)$$

The calculated linear and angular velocities need to be converted into the control action for the motors (right and left wheel motor velocities in this case):

$$v_L = (v_i - \omega_i * d_{base})/r_{wheel} \quad (12.1)$$

$$v_R = (v_i + \omega_i * d_{base})/r_{wheel} \quad (12.2)$$

where $d_{base}$ is the base diameter of the DDMR and $r_{wheel}$ is the radius of the wheel. For the cost, we used the cost function of Linear Quadratic Regulator [27].

$$J = \sum_{i=0}^{N-1} |X_i - X_i^{ref}|^T Q |X_i - X_i^{ref}| + |U_i - U_i^{ref}|^T R |U_i - U_i^{ref}| \quad (13)$$

where the weights $Q \in \mathbb{R}^{3\times 3}$ and $R \in \mathbb{R}^{2\times 2}$ are positive diagonal matrices.

Depending on the reference state and control values used in (13), a point tracking or a path tracking MPC can be deployed. For point tracking only the next point will be given as reference while for path tracking the next $N$ points will be given keeping the initial point fixed as the robot's state. We used a path tracking MPC for our purpose giving it next $N$ points of the path received from the path planning module. This enables us to assign the bot a reference velocity for the whole trajectory instead of the bot slowing down at every point and then moving to the next as in the case of point tracking.

For an improved estimate of the states, we use the 4th order Runge-Kutta method as shown in (14) instead of Euler discretization.

$$x_{i+1} = x_i + \frac{1}{6}(k_1 + 2k_2 + 3k_3 + k_4) \quad (14.1)$$

$$k_1 = hf(x_i, u_i) \quad (14.2)$$

$$k_2 = hf\left(x_i + \frac{h}{2}, u_i + \frac{k_1}{2}\right) \quad (14.3)$$

$$k_3 = hf\left(x_i + \frac{h}{2}, u_i + \frac{k_2}{2}\right) \quad (14.4)$$

$$k_4 = hf(x_i + h, u_i + k_3) \quad (14.5)$$

where $h$ is the step horizon

*E. Adaptive Behaviour*

*1)* Weights in MPC: Classical MPC only allows you to choose a fixed set of weights for your robot. This requires you to over-tune the weights for a specific hardware or environment which might lose its viability if there are changes in the environment or if the code is deployed to another similar bot. Adaptive weights help you to get rid of this problem depending on the conditions of adaptive tuning. Choosing different weights according to the path or environment the robot will face, generalizes the MPC codebase to all other similar systems and the programmed environments. This also makes the tuning easier since there is no longer a need to over-tune the system. As an example, the weight on linear velocity($R_{11}$, where $R_{ij}$ represents the element at $i^{th}$ row and $j^{th}$ column in the R matrix) can be reduced if you have a straight path ahead while the weight on yaw($Q_{33}$) could be increased or equivalently the weight on angular velocity actuation ($R_{22}$) can be decreased if there's a sharp turn ahead.

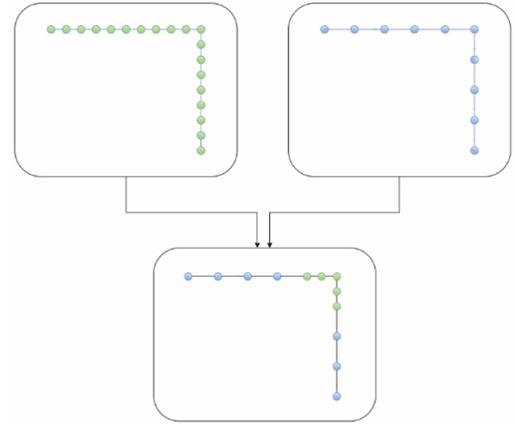

Fig. 3. Top Left: Path for a higher and fixed resolution, Top Right: Path for a lower and fixed resolution and Bottom: Path for adaptive resolution – higher on turns and lower on straight segments.

We initialize the MPC problem in variable format offline to save compuation. So, to change the weights online we append the weights in a paramater matrix which was storing only the reference state and control values before. These parameters are then substituted in the MPC control problem and passed to CasADi to find the optimal solution.

*2)* Resolution for Path Planning: We keep a higher resolution during turns as shown in Fig. 3 so as to ensure accuracy, while the resolution is reduced for straight segments. This allows to save computation as we now have less way points to consider in the optimization problem without the loss of any valuable information. This is because a straight path can be travelled even with way points that are more sparse by making sure the yaw value is maintained which can be done by keeping a higher weight for the yaw in our cost function. This also increases the average linear speed as the controller considers less points to track and hence has to reduce its velocity on number of occasions to ensure convergence on all intermediate targets.

*3)* Corrective Turns: Paths in grid worlds often include sharp turns. A path tracking MPC, since it has to track the next N way points instead of just the current way point, takes a smooth turn at such way points. Since there might be obstacles around the turn, over smoothed paths might cause collisions. Although behaviours like in place turns can be implemented by over- tuning or changing the cost function, this is harder and would make the controller unstable in other situations like straight or diagonal paths. Initially we also used a point tracking for in place turns, but this had other problems as mentioned in the Section IV-D5. In path tracking MPC a high cost on positions in Q matrix was used to implement in place rotations. Although this avoided the obstacles, the behavior was more jerky and took longer times at turns which is undesirable. So we developed a corrective turning algorithm. In this, if the distance between the bot's position and the upcoming turning point is larger than a threshold, then before sending the path to the controller, the path planning module applies a filter. This filter removes the local waypoints after the upcoming turning point. This is continued until the bot is within the threshold upon which the filter is removed. The threshold distance can be easily modelled based on the size of the obstacles. This algorithm along with adaptive weights produced very smooth and quick behaviours. The difference between the classical approach and our approach can be seen in Fig. 4. For an orthogonal path generated by a discrete planner (yellow-dotted line), an initial coarse path is output by the path tracker (red line). Incorporating the turn correction and adaptive weights smoothens the path to give the final trajectory (blue line).

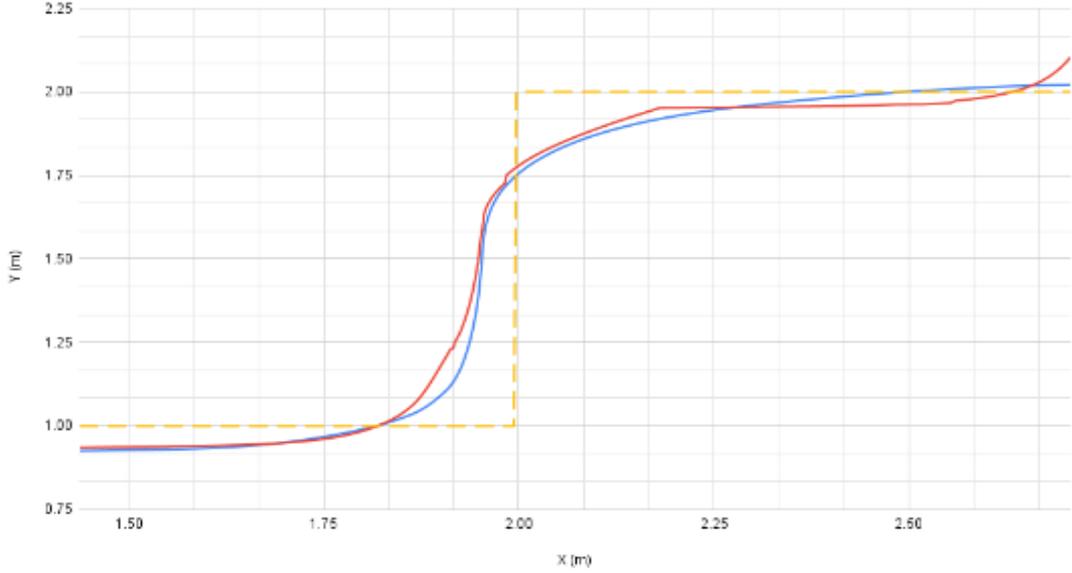

Fig. 4. Trajectory taken by the robot when the turn correction and adaptive weights are activated (blue) and deactivated (red), along with the reference trajectory (yellow) and same for both cases.

## V. EXPERIMENTS

### A. Simulation Setup

We have tested our algorithms in simulation similar to Fig. 5 using Webots [28] which is an open-source simulator for robots. We use TIAGo Base [29] to showcase our differential drive kinematic model for motion control. The simulation has been run on the following - i5 10500H processor, 16 GB RAM and NVIDIA GeForce GTX 1650 graphic drivers.

### B. Experiment 1 - Classical vs Adaptive : Path Planning and Motion Control

To validate our new features (adaptive resolution, adaptive weights and turn correction) added to the path planning and controls module, we use the following metrics in table I:-

- Cross Track Error (CTE): This is the defined as the minimum distance of the robot from the reference path at particular time.

- Jerk: This is defined as change in acceleration and is a metric used in the past [10], [30] to measure smooth motion of the robot. We report linear jerk and angular jerk separately for better understanding of the metric. Using [10], we report $J_{Lin}$ and $J_{Ang}$ which are the average linear and angular jerk respectively for the trajectory and are defined as –

$$J_{Lin} = \frac{\sum_{t=2}^{N-1}|\ddot{v}(t)|*\delta T}{N-2} \quad (15.1)$$

$$J_{Ang} = \frac{\sum_{t=2}^{N-1}|\ddot{\omega}(t)|*\delta T}{N-2} \quad (15.2)$$

where, $t$: Current time step
$N$: Total number of time steps in the trajectory
$\delta T$: Time interval between two consecutive time steps
$\ddot{v}(t)$: Double derivative of linear velocity at t
$\ddot{\omega}(t)$: Double derivative of angular velocity at t

We have used central difference method as shown in (16) to calculate $\ddot{v}(t)$ and $\ddot{\omega}(t)$.

$$\ddot{f}(t_i) = \frac{f_{t+1}-2f_t+f_{t-1}}{\delta T} \quad (16)$$

Here $f_i$ represents value of $f$ at $i^{th}$ time step where $f$ is an arbitrary function.

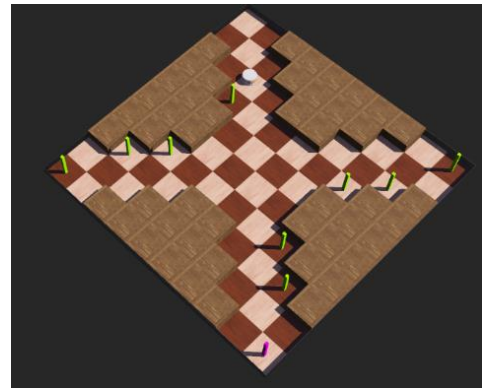

Fig. 5. Sample Webots world in with obstacles, TIAGo bot and 10 waypoints

- Total Traversal Time: This is defined as the time taken for the robot to navigate all waypoints starting from a fixed starting point
- Computational Frequency (Hz) - We measure the average computation frequency of end-to-end framework with all modules included

waypoints was chosen first and then for each case 10 grid maps were generated with number of obstacles chosen randomly between 25 and 35. The positions of obstacles and intermediate waypoints was also chosen randomly while making sure that all waypoints were traversable. For experimentation purposes, the robot was assumed to always start from the (0,0) with the final goal at the (9,9) block of a 10X10 grid which is the top left and

TABLE I. EVALUATION RESULTS FOR COMBINATIONS OF ADAPTIVE RESOLUTION, WEIGHTS AND TURN CORRECTNESS

| Adaptive Resolution | Turn Correction | Adaptive Weights | Cross Track Error (m) | Linear Jerk (m s-2) | Angular Jerk(rad s-2) | Total Traversal Time (s) |
|---|---|---|---|---|---|---|
| × | × | × | 0.11 | **1.728** | 16.758 | 171.162 |
| × | × | ✓ | 0.090 | 2.236 | **1.260** | 181.255 |
| × | ✓ | × | 0.071 | 10.545 | 11.361 | 137.948 |
| ✓ | × | × | 0.098 | 10.717 | 10.557 | 143.788 |
| × | ✓ | ✓ | **0.059** | 3.893 | 2.243 | 138.790 |
| ✓ | × | ✓ | 0.090 | 2.889 | 1.587 | 174.889 |
| ✓ | ✓ | × | 0.106 | 3.393 | 19.357 | **134.341** |
| ✓ | ✓ | ✓ | 0.007 | 5.390 | 3.842 | 152.224 |

TABLE II. ACCURACY AND AVERAGE COMPUTATION TIME OF GREEDY AND PROBABILISTIC METHOD WITH RESPECT TO BCP.

| Number of Waypoints | BCP | | Probabilistic | | Greedy | |
|---|---|---|---|---|---|---|
| | Accuracy (%) | Avg Computation Time(s) | Accuracy (%) | Avg Computation Time(s) | Accuracy (%) | Avg Computation Time(s) |
| 8 | 100 | 0.052 | 82 | 0.0486 | 10 | 0.005 |
| 9 | 100 | 0.203 | 92 | 0.130 | 20 | 0.013 |
| 10 | 100 | 1.150 | 93 | 0.650 | 10 | 0.065 |
| 11 | 100 | 11.382 | 81 | 5.833 | 20 | 0.583 |
| 12 | 100 | 31.125 | 87 | 16.778 | 30 | 1.678 |

The RMS value for Cross Track Error is reported in the table. All the experiments have been done on 10 worlds and the averaged values have been reported in Table I.

### C. Experiment 2 - Evaluating Approaches to find Next Best Waypoint

We define accuracy of methods which were explained in Section IV-B as the accuracy with which it can achieve the best cost which is the cost of the path calculated by the BCP as explained in Section IV-B2. As expected, this means that accuracy of the BCP is 100%. The accuracy of the methods also depends on total number of waypoints. We evaluate all our experiments on the cases where the total number of waypoints vary from 8 to 12. To experiment this, a total number of bottom right blocks respectively in Fig. 1. Now to evaluate all the methods the following experiments were conducted –

- For a fixed $\gamma = 2$, the accuracy and the computation time for all methods are reported in Table II
- Accuracy and computation of probabilistic approach is evaluated relative to BCP while varying $\gamma$ for each case of waypoints. This is shown in Fig. 6.

In both the above experiments the probabilistic approach is run 10 times on each map to find a more robust accuracy.

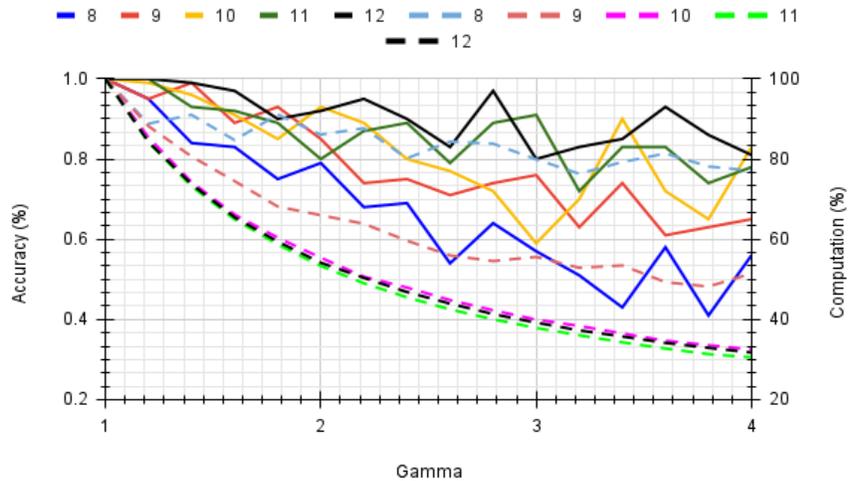

Fig. 6. Accuracy and relative computation of Probabilistic method when compared to BCP at different values of γ and total number of waypoints. Here solid lines represent the accuracy and the dotted lines represent the relative (to γ=1) % of computation used

## VI. RESULTS

As can be seen from Table I, any of our algorithms used alone or with other any other algorithms give better results in term of cross track error when compared to the case where none of the features were used. Adaptive weights having simple conditions on turns and straight paths are shown to perform way better in terms of reducing linear and angular jerks as well as less cross track errors. We can see that the adaptive weights ensure the stability of the system and reduces its wear and tear by reducing the angular jerk by a huge amount. An adaptive resolution which gives less resolution path on straight lines can be seen to increase speed and hence lower time of traversal but with a higher linear jerk due to sparsely placed reference points for straight paths. Turn correction can guarantee no collisions on turns with easier tuning. This along with adaptive weights was our best performing version in terms of CTE and jerks with a time comparable to the least traversal time recorded as in the case of adaptive resolution and adaptive correction. We can see from the Fig. 4, that with adaptive weights algorithm on, the robot is able to take much smoother turns which carry a huge advantage over the jerky turns as in the case of no adaptive weights. Also, the corrective turn algorithm can be seen to delay the turning until a threshold which can be tuned according to the obstacles.

From Table II, we conclude that our novel probabilistic approach provides a variable trade-off between the extremities provided by the optimal yet slow best-cost method and the sub-optimal yet fast greedy planner. We notice that even upon increasing the search-space limiting factor, accuracy of the probabilistic planner for higher number of waypoints is not hurt. This is done by exploiting the fact that in a discretised world, there may exist multiple global trajectories with the same path costs.

We see that as the number of waypoints in the test space is increased, the decrease in accuracy (∆accuracy) decreases, and is also more random. This is due to the fact that as the number of waypoints increase, there exists a higher probability of multiple solutions with the same cost, as can be seen in Fig. 7, which was obtained by calculating the average number of paths (averaged over 40 iterations with a fixed gamma) that have cost equal to the best cost for that γ. This in turn increases the probability of a fractional search space containing near optimal low-cost solutions. This can also be observed from the fact that the accuracy of maps with 8 waypoints give an average accuracy of 80% when the search-space is limited to about 0.7 of its size, while the same accuracy was achieved for maps with 11 and 12 waypoints, even when just 1/4th of the search-space is used. It can also be observed from Fig. 7 that for a fixed number of waypoints while we increase the value of γ, as expected the average number of paths with cost equal to optimal cost decreases. This is intuitively due to the lesser probability of an optimal cost path to be part of the reduced search-space.

From Fig. 6, we also see an exponential decrease in the computation used as γ increases. This is due to the innate nature of the search-space limiting factor, which limits the search-space more if it is large to begin with, which is true in case of permutations of higher number of waypoints. We also observe a plateauing trend in the computation as γ increases. This is due to the trend of the factorial function, which increases very steeply, and thus as higher integral values, even a smaller search-space

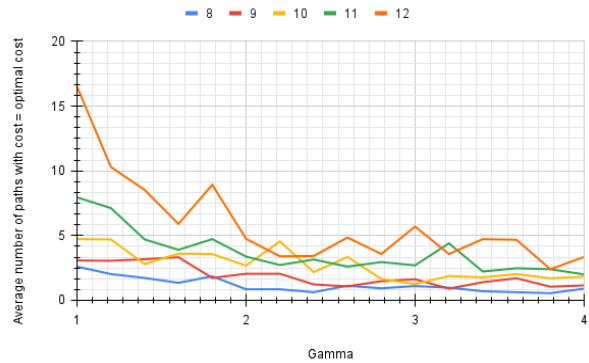

Fig. 7. Average number of paths in the reduced search space for a given g and with cost equal to optimal cost. Color of the graphs represent the total number of waypoint

does not offer computational liberties. The simulations in this work were limited mainly due to this increase in waypoint permutations, which caused a factorial proportional increase in processing requirement. For instance, performing best-cost planning for thirteen waypoints required 5.2 GB of RAM, while for fourteen waypoints required 72.5 GB, limiting the experimentation capacity of this work.

## VII. CONCLUSION

In this paper, we have developed a planning and controls stack for a differential-drive robot, for multiple waypoint navigation in indoor cluttered environments indoor navigation. As part of our work, we introduce a novel probabilistic path planner which provides a computation-optimality trade-off by limiting the search-space of the best-cost planner. We also introduce a simple adaptive weight Model Predictive Control algorithm along with improved turns and adaptive path resolution. Experiments on our framework show a time-efficient online planning ability provided by the planner, coupled with the robust and adaptive control strategy, which together enable the agent to traverse unknown indoor environments in the most efficient way possible.

## VIII. FUTURE WORK

For future work, a good starting point can be hardware implementation. Then, improvements to the cost functions can be done in IV-B by taking into account predicted number of turns to reach a waypoint. Other approaches like kinodynamic planning and diagonal movement in the global path planning can be incorporated as well. To make the control stack more robust, we propose a dynamic model to be used as part of the MPC controller for modelling systems dynamics. Approaches like a reinforcement learning based control strategy can be explored as well.

## ACKNOWLEDGMENT

This work has taken place at the Autonomous Ground Vehicle (AGV) Research Group at Indian Institute of Technology Kharagpur, under Professor Debashish Chakravarty.